\documentclass[sigplan]{acmart}
\renewcommand\footnotetextcopyrightpermission[1]{} 

\acmConference[Radio'21]{VMware Radio conference}{2021}{SF, USA}
\acmYear{2021}
\acmISBN{} 
\acmDOI{} 
\startPage{1}

\setcopyright{none}

\usepackage{booktabs}   
\usepackage{subcaption} 
                        
\usepackage{mathptmx}      
\usepackage[labelfont=bf]{caption}

\usepackage{graphicx}
\usepackage [english]{babel}
\usepackage [autostyle, english = american]{csquotes}
\MakeOuterQuote{"}
\usepackage{listings}
\usepackage{enumerate} 
\usepackage[section]{placeins}

\usepackage{booktabs}
\usepackage{multirow}
\usepackage{float}

\usepackage{hyperref}
\usepackage{mathtools}

\begin{document}

\title[ DriveML]{ DriveML: An R Package for Driverless Machine Learning }  

\author{Sayan Putatunda}
\authornote{Corresponding author}          
\affiliation{
 \position{Senior Manager- AA \& DS}
  \department{Enterprise Data and Analytics (EDA)}              
  \institution{VMware Software India Pvt. Ltd.}            
  \country{Bangalore, India}                    
}
\email{sayanp@iima.ac.in}          

\author{Dayananda Ubrangala}
\affiliation{
  \position{Senior Business Analyst- AA \& DS}
  \department{Enterprise Data and Analytics (EDA)}              
  \institution{VMware Software India Pvt. Ltd.}            
  \country{Bangalore, India}                    
}
\email{daya6489@gmail.com}         

\author{Kiran R}
\affiliation{
  \position{Senior Director- AA \& DS}
  \department{Enterprise Data and Analytics (EDA)}              
  \institution{VMware Software India Pvt. Ltd.}            
  \country{Bangalore, India}                    
}
\email{efpm04013@iiml.ac.in}         

\author{Ravi Prasad Kondapalli}
\affiliation{
  \position{Senior Manager- AA \& DS}
  \department{Enterprise Data and Analytics (EDA)}              
  \institution{VMware Software India Pvt. Ltd.}            
  \country{Bangalore, India}                    
}
\email{rpkondapalli@yahoo.com}         

\renewcommand{\shortauthors}{Putatunda, et al.}
\begin{abstract}
In recent years, the concept of automated machine learning has become very popular. Automated Machine Learning (AutoML) mainly refers to the automated methods for model selection and hyper-parameter optimization of various algorithms such as random forests, gradient boosting, neural networks, etc. In this paper, we introduce a new package i.e. DriveML for automated machine learning. DriveML helps in implementing some of the pillars of an automated machine learning pipeline such as automated data preparation, feature engineering, model building and model explanation by running the function instead of writing lengthy R codes. The DriveML package is available in CRAN. We compare the DriveML package with other relevant packages in CRAN/Github by applying them on multiple datasets of different dimensions. We find that DriveML performs the best taking into consideration both the prediction accuracy and the execution time. We also provide an illustration by applying the DriveML package with default configuration on a real world dataset. Overall, the main benefits of DriveML are in development time savings, reduce developer's errors, optimal tuning of machine learning models and reproducibility.
\end{abstract}



\setcopyright{acmcopyright}
\copyrightyear{2021}
\acmYear{2021}

\acmConference[SIGKDD '21, 2nd International Workshop on
Data Quality Assessment 
for Machine Learning]{2nd International Workshop on
Data Quality Assessment 
for Machine Learning}{August 14--18, 2021}{Singapore}
\acmBooktitle{2nd International Workshop on
Data Quality Assessment 
for Machine Learning, @ Special Interest Group on Knowledge Discovery and Data Mining (SIGKDD), August 14--18, 2021, Singapore}


\maketitle

\section{Introduction} \label{intro}
Machine learning has disrupted almost every industry around us such as social media, transportation, agriculture, retail, software development, marketing, sales, finance, manufacturing and more. Machine learning (ML) is the phenomenon by which computers learn things by themselves and recognize different patterns \citep{ml:1, ml:3}. Thomas M. Mitchell defined the machine learning problem as, \textit{"A computer program is said to learn from experience E with respect to some class of tasks T and performance measure P, if its performance at tasks in T, as measured by P, improves with experience E"} \citep{ml:2}. 

In recent years, the concept of automated machine learning has gained traction. Many of the top tech companies such as Google, Amazon, Facebook, H2O and more (including some tech startups) are focusing on this. Automated Machine Learning (AutoML) mainly refers to the automated methods for model selection and hyper-parameter optimization of various algorithms such as random forests, gradient boosting, neural networks and more \citep{auto:2}. The four pillars of building an automated machine learning pipeline are: (a) Data Preparation, (b) Feature Engineering, (c) Model development and (d) Model evaluation. Any AutoML pipeline would focus on automating these four pillars \citep{auto:1}. Please see Elshawi et al. \citep{auto:3} for a detailed overview of some of the recent works reported in the literature for automated machine learning. 

In this paper, we introduce the DriveML package for automated machine learning especially in the classification context. DriveML saves a lot of effort required for data preparation, feature engineering, model selection and writing lengthy codes in a programming environment such as R \citep{R:11}. Thus, the DriveML package saves time and leads to more efficiency. In the following sections, we discuss the key functionality of the package, provide an illustration by applying the DriveML package with default configuration on a real world dataset, and compare DriveML with other relevant R packages across multiple datasets.

\section{Key Functionality} \label{func}
\begin{figure}[!htp]
 \centering
  \includegraphics[width=0.5\textwidth]{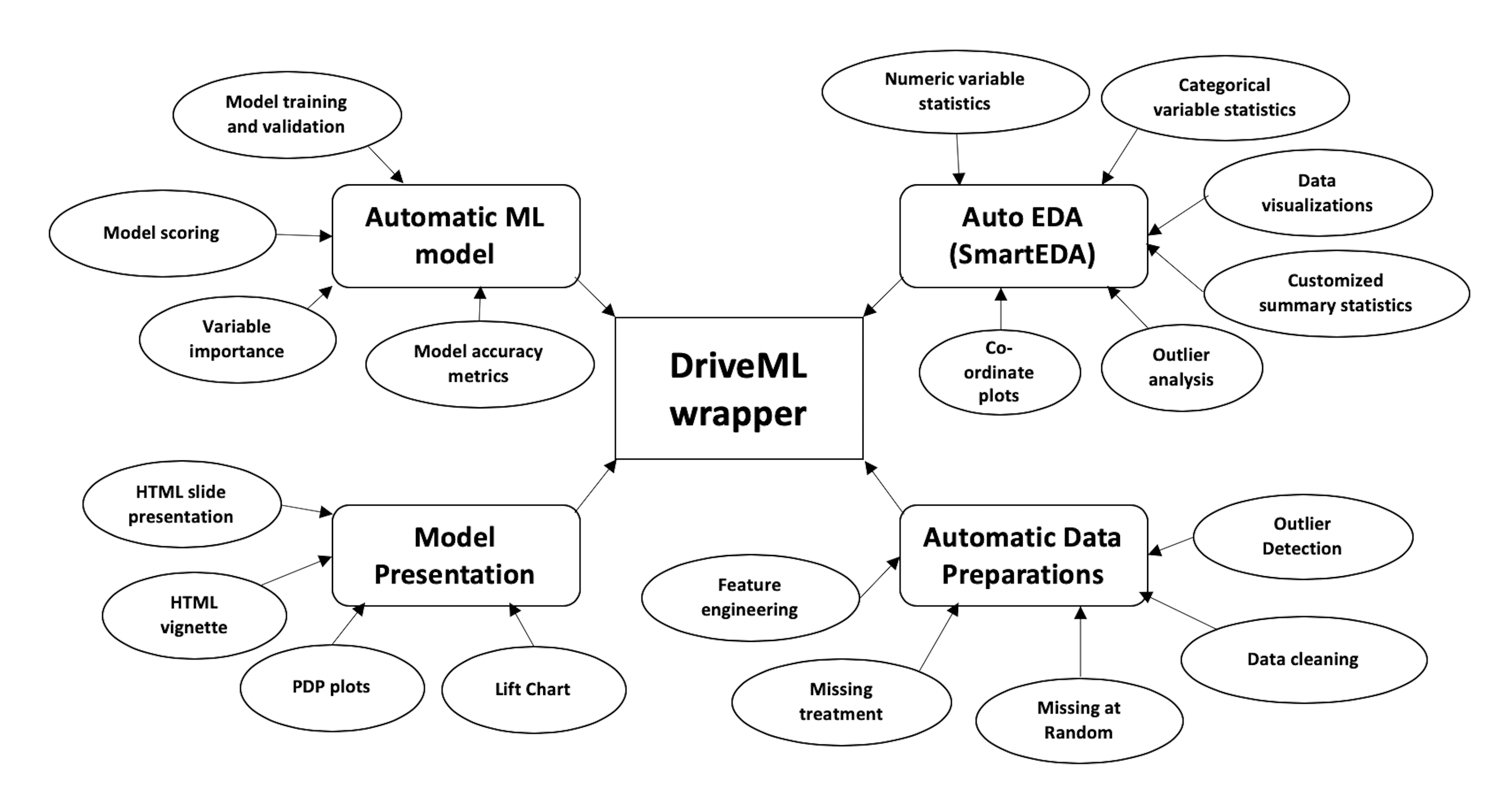}
  \caption{The various functionalities of DriveML.}
  \label{fig:boat1}
\end{figure}

Figure \ref{fig:boat1} shows the various functionalities of the DriveML package. DriveML has a single function i.e. "autoDataprep" that performs automatic data preparation steps on the raw input dataset. The various steps involve (a) Data cleaning- NA/infinite values are replaced, duplicates are removed and feature names are cleaned, (b) Missing value treatment- such as mean/median imputation or imputation using the MLR package \citep{R:4}, (c) Outlier detection- creating an outlier flag and imputation with 5th or 95th percentile value  and (d) Feature engineering- it performs various operations such as missing at random features, Date variable transformation, bulk interactions for numerical features, one-hot encoding for categorical variables and finally, feature selection using zero variance, correlation and Area under the curve (AUC) method. 

DriveML can call the SmartEDA \citep{smarteda:1} package for performing automated exploratory data analysis (EDA). The "autoMLmodel" function in DriveML performs the various tasks for automated machine learning such as creating model test and train datasets  and  then run multiple classification models such as 
(i) glmnet- Regularised regression from glmnet R package \citep{R:10}, (ii) logreg- logistic regression from stats R package \citep{R:11}, (iii) randomForest- Random forests using the randomForest R \citep{R:12}, (iv) ranger- Random forests using the ranger R package \citep{R:13}, (v) xgboost- Extreme Gradient boosting using xgboost R package \citep{R:14} and (vi) rpart- decision tree classification using rpart R package \citep{R:15}.

The other features of the autoMLmodel function are (a) Hyper-parameter Tuning- using Random search (default option) or by applying the irace (Iterated Racing for Automatic Algorithm Configuration) package \citep{irace:1} , (b) performs model validation using Receiver Operating Characteristics Area Under the Curve (ROC AUC), (c) Model scoring and (d) Variable importance chart for the best model (i.e. the one having the highest ROC AUC in the validation dataset). 

The autoMLmodel function will also output the lift chart for the best model and will also provide the Partial dependency plot (PDP plots) for the top five variables from the best model. If the user wants to get the PDP plots for some specific variables, then they can user the function "autoPDP" for the same. 
The "autoMAR" function is used to identify and generate the Missing at Random features (MAR). The following steps explain the way it works. (1) For every feature X1 in a dataset with missing values, we create a new variable Y1, which will have value of 1 if X1 has a missing value or 0 otherwise,  (2) We then fit a classification model with Y1 as dependent variable and all other features other than X1 as independent variables, (3) If the AUC is high, it means that there is a pattern to the missing values( i.e. they are not missing at random), in this case we retain Y1 as an additional independent variable in the original dataset, 
(4) If the AUC is low, then the missing values in X1 are missing at random, then Y1 is dropped, (5) Repeat steps 1 to 4 for all the independent variables in the original dataset  and (6) Publish a report with the findings for each independent variable X and additional variables added to the dataset.

Finally, DriveML will provide an HTML slide/vignette report containing the data descriptive statistics, model results, AUC plots, lift charts and PDP plots.

\section{Illustration} \label{results}
\begin{figure}[!htp]
 \centering
  \includegraphics[width=0.5\textwidth]{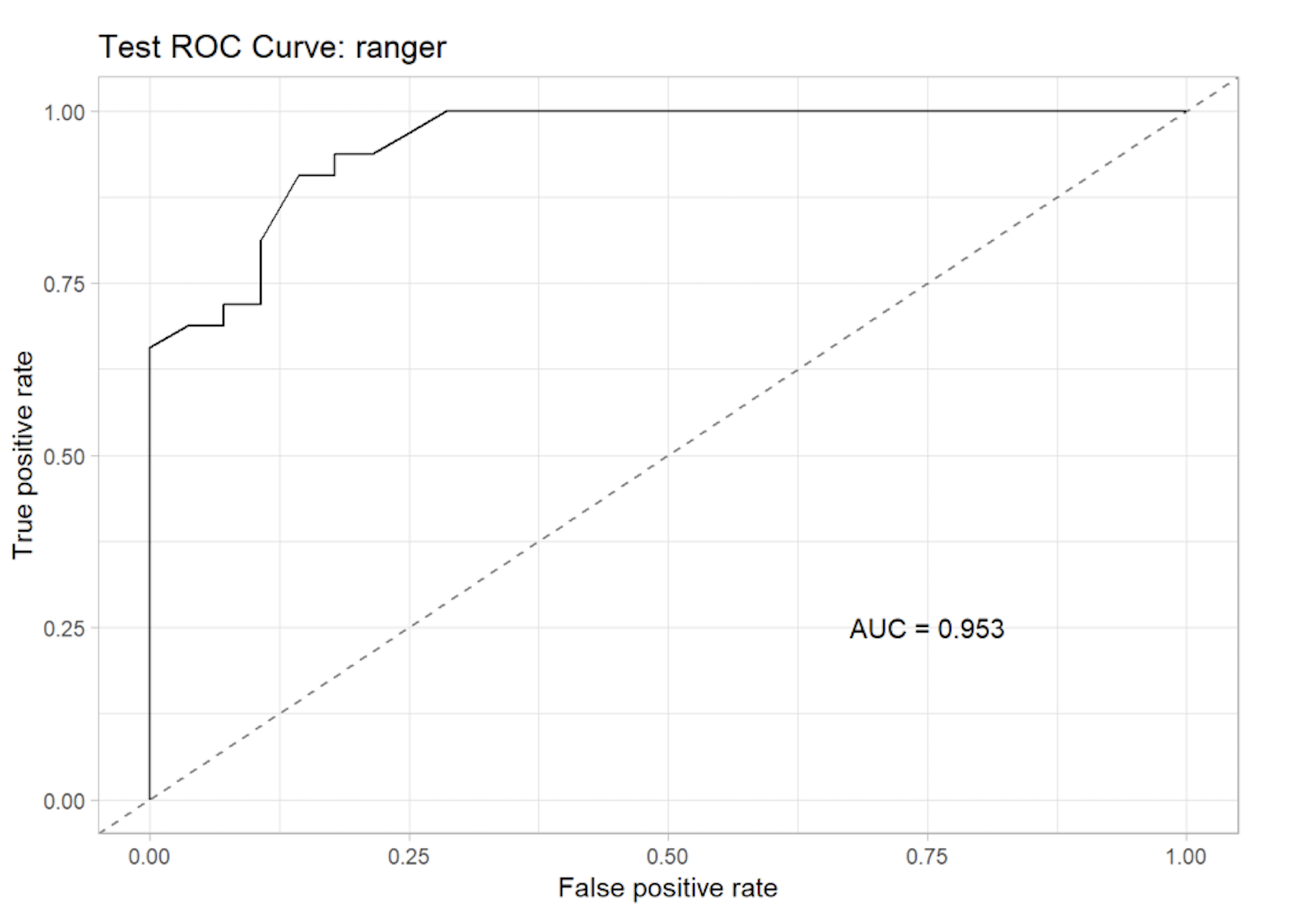}
  \caption{Test AUC plot for the ranger method.}
  \label{fig:boat4}
\end{figure}

In this section, we provide an illustration by applying the DriveML package with default configuration on a real world dataset. The goal is to make an user of the DriveML package familiar with the various functions (please see Section 2) of the package in R. We use the Heart disease dataset that is publicly available in the UCI machine learning repository \citep{R:9}. The dataset has 303 observations and 14 variables. The target variable is a binary categorical variable that indicates whether a patient has heart disease or not. The independent variables are age, gender, presence of chest pain, cholesterol, resting blood pressure, maximum heart rate and more. We apply the DriveML package and it splits the dataset into 243 observations in train and 60 observations in test datasets. The results for the various methods are shown below in Table \ref{res21}. We find that the ranger method gives the best result in terms of the Test AUC. We consider the Test AUC as the default evaluation metric in this package. However, the user is free to choose any other evaluation metrics such as (F1 score, Precision, Recall and Accuracy) as well. We also record the model fitting time and the scoring time as shown in Table \ref{res21}. Figure \ref{fig:boat4} shows the test AUC plot. The modus-operandi of applying DriveML in R is described in the code snippet below.

We first load the dataset and perform data preparation using the "autoDataprep" function.
\begin{verbatim}
> library("DriveML")
> library("SmartEDA")

## Load the dataset 
> heart = DriveML::heart

#Now, let us perform data preparation using DriveML.
> dateprep <- autoDataprep(data = heart,
                         target = 'target_var',
                         missimpute = 'default',
                         auto_mar = FALSE,
                         mar_object = NULL,
                         dummyvar = TRUE,
                         char_var_limit = 15,
                         aucv = 0.002,
                         corr = 0.98,
                         outlier_flag = TRUE,
                         uid = NULL,
                         onlykeep = NULL,
                         drop = NULL)

> printautoDataprep(dateprep)
\end{verbatim}

Now, we will perform automated training, tuning, and validation of the different machine learning models using the "autoMLmodel" function. This function includes six binary classification techniques that were mentioned earlier. We then generate the model summary results as shown in Table \ref{res21} and plot the Test ROC AUC plot.

\begin{verbatim}
> mymodel <- autoMLmodel( train = heart,
                        test = NULL,
                        target = 'target_var',
                        testSplit = 0.2,
                        tuneIters = 10,
                        tuneType = "random",
                        models = "all",
                        varImp = 10,
                        liftGroup = 50,
                        maxObs = 4000,
                        uid = NULL,
                        htmlreport = FALSE,
                        seed = 1991)
                        
# Model summary results
> mymodel$results

# Plot the Test ROC AUC
> TestROC <- 
mymodel$trainedModels$randomForest$modelPlots$TestROC
> TestROC
\end{verbatim}

\begin{table*}[htp]
\centering
\caption{Performance comparison of different techniques on the Heart disease dataset}
\label{res21}
 \scalebox{1}{
\begin{tabular}{ccccccccc}
\hline 
Model        & \begin{tabular}[c]{@{}c@{}}Fitting time \\ (secs)\end{tabular} & \begin{tabular}[c]{@{}c@{}}Scoring time \\ (secs)\end{tabular} & Train AUC & Test AUC & Accuracy & Precision & Recall & F1\_score \\ \hline 
ranger       & 3.148                                                          & 0.017                                                          & 0.997     & 0.953    & 0.867    & 0.816     & 0.969  & 0.886     \\\hline 
glmnet       & 3.377                                                          & 0.009                                                          & 0.915     & 0.941    & 0.867    & 0.833     & 0.938  & 0.882     \\\hline 
logreg       & 4.384                                                          & 0.005                                                          & 0.915     & 0.940     & 0.867    & 0.833     & 0.938  & 0.882     \\\hline 
randomForest & 3.155                                                          & 0.011                                                          & 0.997     & 0.937    & 0.850     & 0.811     & 0.938  & 0.870      \\\hline 
xgboost      & 3.56                                                           & 0.005                                                          & 0.996     & 0.930     & 0.867    & 0.816     & 0.969  & 0.886     \\\hline 
rpart        & 2.799                                                          & 0.005                                                          & 0.908     & 0.859    & 0.833    & 0.806     & 0.906  & 0.853    \\ \hline 
\end{tabular}}
\end{table*}

Finally, we can use the "autoMLReport" function to generate a report in the html format for the output of "autoDataprep" and "autoMLmodel" DriveML functions.
\begin{verbatim}
> autoMLReport(mlobject = mymodel, mldata = heart, 
     op_file = "driveML_ouput_heart_data.html")
\end{verbatim}

\begin{figure}[!htp]
 \centering
  \includegraphics[width=0.5\textwidth]{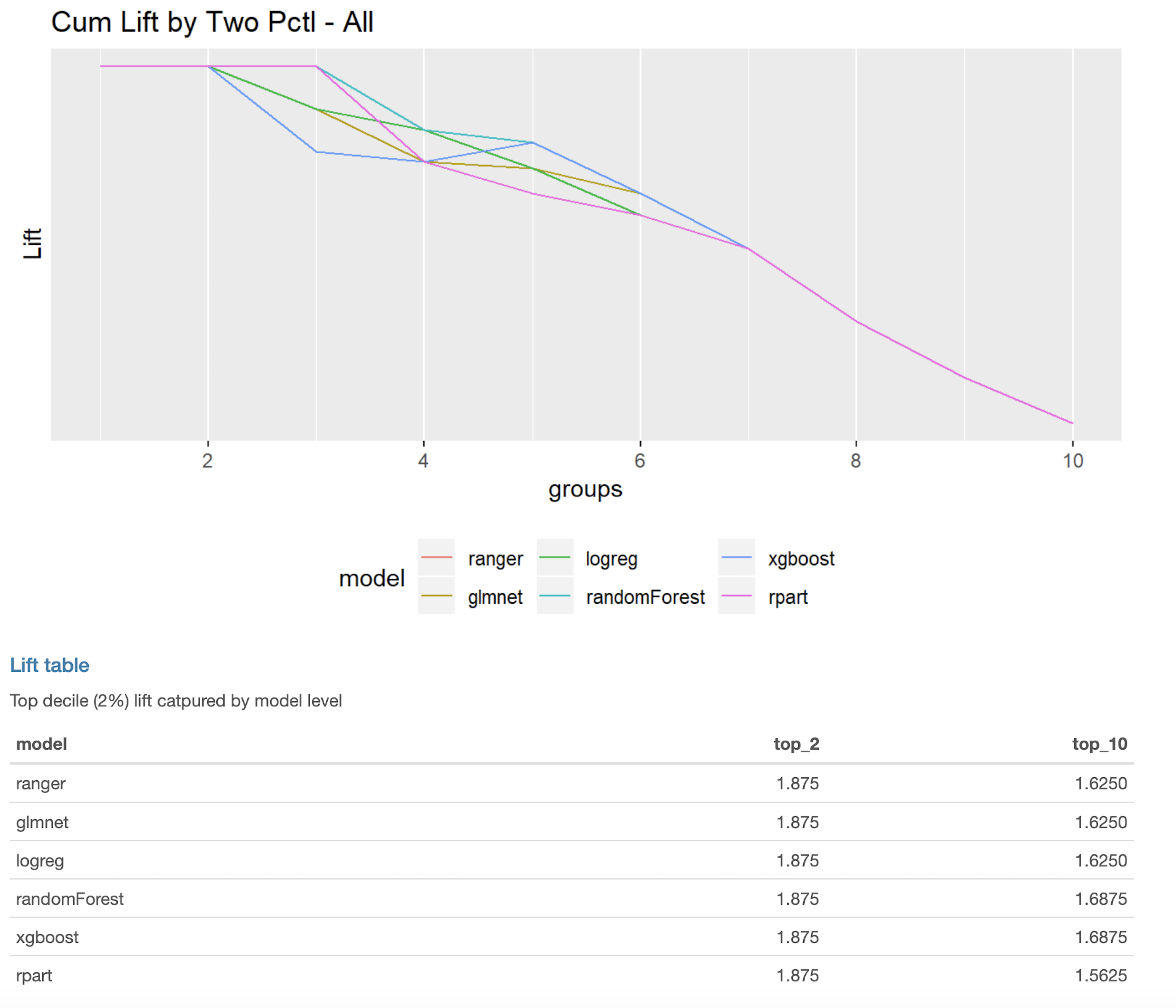}
  \caption{Lift chart and lift table for all the six methods.}
  \label{fig:boat5}
\end{figure}

\begin{figure}[!htp]
 \centering
  \includegraphics[width=0.5\textwidth]{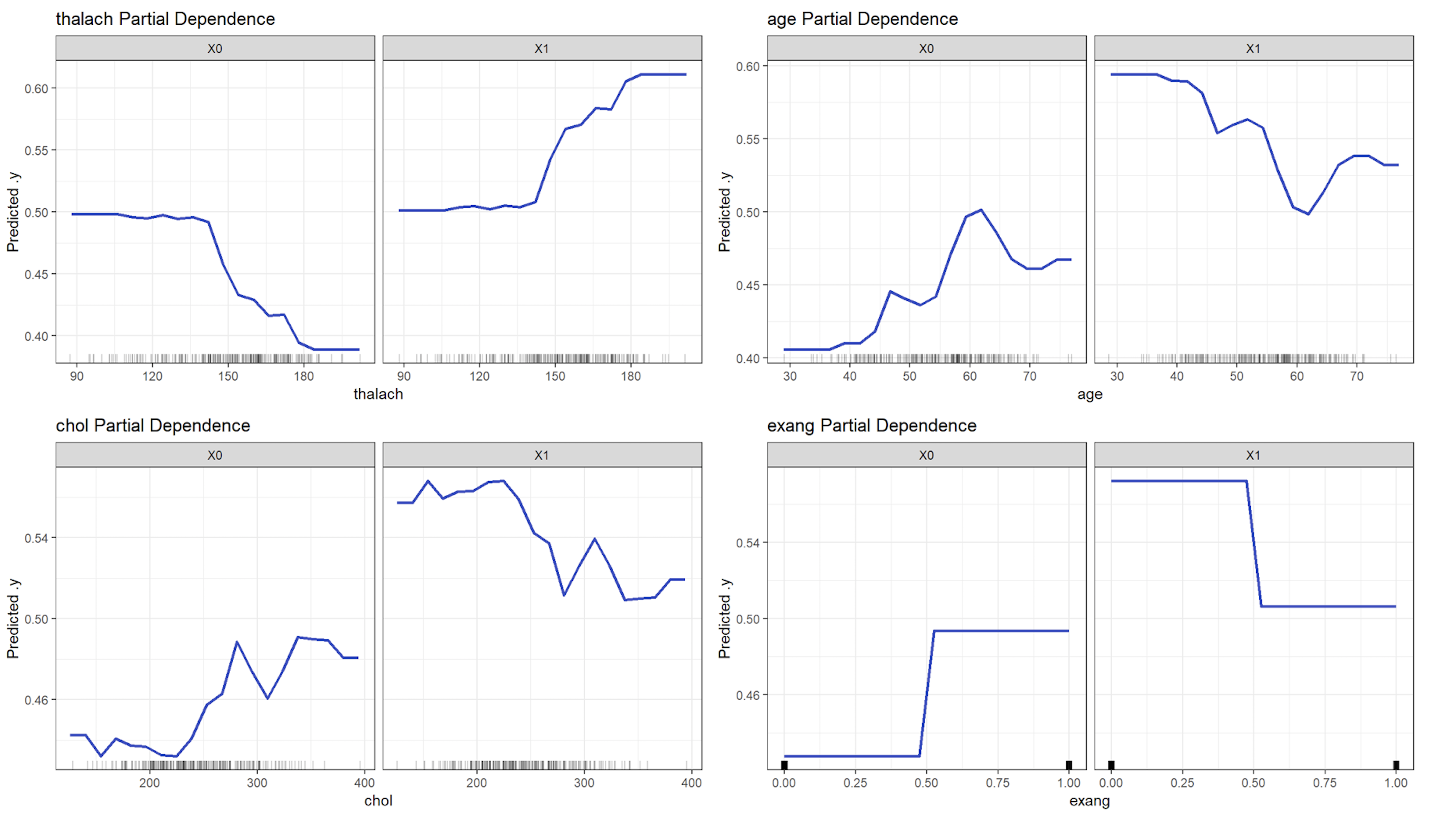}
  \caption{Partial Dependency plots (PDP) for variables- age, thalach, chol and exang.}
  \label{fig:boat6}
\end{figure}

Figures \ref{fig:boat5} and \ref{fig:boat6} show the lift chart for all the methods (along with the lift tables) and the partial dependency plots (PDP) for some of the independent variables respectively.

Moreover, we applied the DriveML package on much larger datasets and we got some encouraging results taking into account both the prediction accuracy and the time taken (we will discuss these in more details in Section \ref{comp}). However, the html format reports for each of these experiments that demonstrates the output of the DriveML functions are available in the "Articles" section of the Github webpage of the DriveML package: \url{https://daya6489.github.io/DriveML/}. 

\section{Comparison of DriveML with other relevant R Packages} \label{comp}
In this section, we compare the DriveML package with other similar packages available in CRAN/Github for automated machine learning namely, OneR \citep{R:5}, H2O \citep{R:6}, and AutoXGBoost \citep{autoxgb:1}. We will apply these packages on four different datasets of different dimensions in the context of binary classification. The metrics for evaluation would be prediction accuracy (i.e. the Test AUC in this case) and the total execution time in seconds (this includes features engineering and model training).

However, please note that there are some limitations that some of these competing packages have when compared to that of DriveML. For example, the AutoXGBoost function only used the Extreme Gradient Boosting (XGBoost) method for automated machine learning. Also, both OneR and AutoXGBoost don't possess the capability for automated feature engineering, missing value treatment, and more. So, to ensure an apples to apples comparison, we will only use Boosting algorithms for both DriveML and H2O autoML functions. And we will also include the dataset preparation time in the total execution time metric for both OneR and AutoXGBoost in the experiments conducted by us.

We use datasets of different dimensions/sizes and categorize them into four different categories such as Extra large, Large, Medium, and Small. The extra large one is the rain in Australia dataset (source: \url{https://www.kaggle.com/jsphyg/weather-dataset-rattle-package}). This dataset contains about 10 years of daily weather observations from many locations across Australia. It contains the target variable "RainTomorrow", which signifies whether it will rain tomorrow or not. The total number of observations in the training dataset are $116,368$. We will refer to this dataset as  "Weather Australia" for the rest of this paper. Moreover, this dataset also contains missing values.

The next dataset is the "Adult data" and we classify this into the Large dataset size category. This is a census data and is publicly available in the UCI machine learning repository \citep{R:9}. The binary target variable indicates if an individual's income exceeds $50,000$ USD per year. The dataset that we consider in the medium size category is the "HR Analytics" dataset that is available in Kaggle (source: \url{https://www.kaggle.com/arashnic/hr-analytics-job-change-of-data-scientists}). This dataset is designed by HR researchers to understand the factors that lead a person to leave a current job. And finally, we use the Heart disease dataset that is classified as a small dataset size category. It is publicly available in the UCI machine learning repository \citep{R:9}. All the information regarding number of attributes, if there are missing values, number of training and test dataset observations for each of these four datasets are provided in Table \ref{res31}.

\begin{table*}[htp]
\centering
\caption{Comparison of DriveML with other available R packages}
\label{res31}
 \scalebox{0.9}{
\begin{tabular}{cccccccccc}
\hline
Sl no. & \begin{tabular}[c]{@{}c@{}}Dataset \\ name\end{tabular}                       & \begin{tabular}[c]{@{}c@{}}Dataset \\ size category\end{tabular} & R package   & \begin{tabular}[c]{@{}c@{}}No. of \\ Attributes\end{tabular} & \begin{tabular}[c]{@{}c@{}}Missing \\ values?\end{tabular} & \begin{tabular}[c]{@{}c@{}}No. of Train \\ instances\end{tabular} & \begin{tabular}[c]{@{}c@{}}No. of Test \\ instances\end{tabular} & \begin{tabular}[c]{@{}c@{}}Total Execution \\ time (in secs)\end{tabular} & Test AUC \\ \hline
1      & \multirow{4}{*}{\begin{tabular}[c]{@{}c@{}}Weather \\ Australia\end{tabular}} & \multirow{4}{*}{Extra Large}                                     & DriveML     & \multirow{4}{*}{23}                                          & \multirow{4}{*}{yes}                                       & \multirow{4}{*}{116368}                                           & \multirow{4}{*}{29092}                                           & 263.29                                                                    & 0.89     \\
2      &                                                                               &                                                                  & H2O automl  &                                                              &                                                            &                                                                   &                                                                  & 1176.91                                                                   & 0.89     \\
3      &                                                                               &                                                                  & OneR        &                                                              &                                                            &                                                                   &                                                                  & 1.19                                                                      & 0.77     \\
4      &                                                                               &                                                                  & autoxgboost &                                                              &                                                            &                                                                   &                                                                  & 139.16                                                                    & 0.88     \\ \hline
5      & \multirow{4}{*}{\begin{tabular}[c]{@{}c@{}}Adult \\ data\end{tabular}}        & \multirow{4}{*}{Large}                                           & DriveML     & \multirow{4}{*}{14}                                          & \multirow{4}{*}{No}                                        & \multirow{4}{*}{32561}                                            & \multirow{4}{*}{16281}                                           & 29.90                                                                     & 0.92     \\
6      &                                                                               &                                                                  & H2O automl  &                                                              &                                                            &                                                                   &                                                                  & 47.84                                                                     & 0.92     \\
7      &                                                                               &                                                                  & OneR        &                                                              &                                                            &                                                                   &                                                                  & 0.13                                                                      & 0.71     \\
8      &                                                                               &                                                                  & autoxgboost &                                                              &                                                            &                                                                   &                                                                  & 40.70                                                                     & 0.92     \\ \hline
9      & \multirow{4}{*}{\begin{tabular}[c]{@{}c@{}}HR \\ Analytics\end{tabular}}      & \multirow{4}{*}{Medium}                                          & DriveML     & \multirow{4}{*}{14}                                          & \multirow{4}{*}{yes}                                       & \multirow{4}{*}{15327}                                            & \multirow{4}{*}{3831}                                            & 35.13                                                                     & 0.79     \\
10     &                                                                               &                                                                  & H2O automl  &                                                              &                                                            &                                                                   &                                                                  & 60.87                                                                     & 0.80     \\
11     &                                                                               &                                                                  & OneR        &                                                              &                                                            &                                                                   &                                                                  & 0.11                                                                      & 0.71     \\
12     &                                                                               &                                                                  & autoxgboost &                                                              &                                                            &                                                                   &                                                                  & 13.55                                                                     & 0.64     \\ \hline
13     & \multirow{4}{*}{\begin{tabular}[c]{@{}c@{}}Heart\\  Disease\end{tabular}}     & \multirow{4}{*}{Small}                                           & DriveML     & \multirow{4}{*}{14}                                          & \multirow{4}{*}{No}                                        & \multirow{4}{*}{243}                                              & \multirow{4}{*}{60}                                              & 6.07                                                                      & 0.91     \\
14     &                                                                               &                                                                  & H2O automl  &                                                              &                                                            &                                                                   &                                                                  & 32.07                                                                     & 0.90     \\
15     &                                                                               &                                                                  & OneR        &                                                              &                                                            &                                                                   &                                                                  & 0.05                                                                      & 0.68     \\
16     &                                                                               &                                                                  & autoxgboost &                                                              &                                                            &                                                                   &                                                                  & 10.29                                                                     & 0.90    \\ \hline
\end{tabular}}
\end{table*}

Table \ref{res31} describes the results of the experiments conducted by us to compare the performance of DriveML against other R packages across different datasets as mentioned earlier. All the experiments were conducted on a machine with configuration: 32 GB RAM, 64-bit Windows OS, intel i7 processor, and CPU @ 2.60GHz. In Table \ref{res31}, we can see that for a small dataset i.e. the Heart disease dataset, the DriveML is claerly the best performer taking into account both test AUC and the time taken. However, for the other datasets, the test AUC scores for both DriveML and H2O autoML are very close. But DriveML is much faster than H2o autoML. Thus, we find that DriveML is the best performer taking into consideration both prediction accuracy and total execution time taken.

\section{Conclusion} \label{conclusion}
The contribution of this paper is in the development of a new package in R i.e., DriveML for automated machine learning. DriveML helps in implementing some of the pillars of an automated machine learning pipeline such as automated data preparation, feature engineering, model building (techniques such as Random forest, XGBoost, logistic regression and more) and model explanation (using lift chart and PDP plots) by running the function instead of writing lengthy R codes. DriveML also provides some additional features such as model ensembling, lift charts and automated exploratory data analysis when compared to other R packages. Moreover, DriveML also exports the model results with the required plots in an HTML vignette report format that follows the best practices of the industry and the academia. Overall, the main benefits of the DriveML package are in development time savings, reduce developer's errors, optimal tuning of machine learning models and reproducibility.

\section*{Availability}
The software is distributed under an MIT + file LICENSE (Repository: CRAN) and is available from \url{https://github.com/daya6489/DriveML}.

\section*{Acknowledgments}
We would like to thank VMware and the Enterprise \& Data Analytics (EDA) leadership for giving us the required infrastructure and support for this work. We are grateful to the R community for their acceptance and feedback to improve our package further.

%


\bibliography{paper}   




\end{document}